\documentclass[final,3p,12pt,times,authoryear]{elsarticle}

\usepackage{amssymb}
\usepackage{algorithm}
\usepackage{algpseudocode}
\usepackage{amsmath}
\usepackage{enumitem}

\usepackage{booktabs}
\usepackage{array}
\usepackage{graphicx}
\usepackage{multirow}
\usepackage{makecell}
\usepackage{xcolor}
\usepackage{tabularx}
\usepackage{hyperref}
\usepackage{subcaption}

\definecolor{darkblue}{RGB}{0,30,139}
\definecolor{darkteal}{RGB}{0,128,128}

\hypersetup{
  colorlinks=true,
  linkcolor=darkblue,
  citecolor=darkblue,
  urlcolor=darkblue,
  pdfborder={0 0 0}
}

\begin{document}

\begin{frontmatter}

\title{Robust Ranking of Port Impacts from Impending Tropical Cyclones: A Two-Stage Recommendation System}

\author[inst1]{Chenchen Kuai}

\author[inst2]{Shixuan Tang}

\author[inst1]{Chaolun Ma}

\author[inst3]{Xiaoqiang Kong}

\author[inst1]{Xiubin Bruce Wang}

\author[inst1]{\protect\\ Yunlong Zhang}

\author[inst2]{Zihao Li\corref{cor1}}

\ead{zihao.li@marquette.edu}

\affiliation[inst1]{
  organization={Zachry Department of Civil \& Environmental Engineering,
  Texas A\&M University},
  addressline={3136 TAMU},
  city={College Station},
  state={TX},
  postcode={77843},
  country={USA}
}

\affiliation[inst2]{
  organization={Department of Civil, Construction and Environmental Engineering,
  Marquette University},
  addressline={1515 West Wisconsin Avenue},
  city={Milwaukee},
  state={WI},
  postcode={53233},
  country={USA}
}

\affiliation[inst3]{
  organization={Texas A\&M Transportation Institute},
  addressline={505 East Huntland Drive},
  city={Austin},
  state={TX},
  country={USA}
}

\cortext[cor1]{Corresponding author: Zihao Li.}

\begin{abstract}

When a tropical cyclone (TC) threatens several ports, preparedness planning depends on anticipating which ports are likely to be affected and where disruptions may last longest. Limited historical observations at individual ports and uncertainty in landfall location make these relative impacts difficult to estimate. We use a two-stage recommendation framework that first selects candidate ports based on their proximity to the expected landfall location and then ranks them by their estimated number of impact days. Using AIS-derived impact days for 37 ports and 31 TCs in the Gulf of Mexico study region, we compare three ways of representing port--cyclone interactions: pairwise feature interactions, tree-based feature combinations, and explicit feature crosses combined with learned embeddings. In tests on Hurricanes Nicole and Ian, the retrieval stage consistently identifies all observed affected ports under landfall uncertainty. Among the ranking models, Wide \& Deep Learning achieves the closest agreement with the observed port rankings. Its mean ranking agreement exceeds that of the distance-based baseline by 85.5\% for Nicole and 55.5\% for Ian and is less sensitive to uncertainty in the landfall location. These rankings could help port authorities prioritize post-cyclone inspections and allocate recovery resources across ports threatened by the same event.

\end{abstract}


\end{frontmatter}

\section{Introduction}

Maritime transportation plays a central role in U.S. international trade. In 2016, water transportation accounted for approximately 69\% of U.S. imports and exports by weight and 40\% by value \citep{BTS2017FreightFacts}. Ports connect these maritime flows with regional road, rail, and inland waterway networks \citep{ng2015climate}. Their coastal locations also expose them to tropical cyclones (TCs), which can damage infrastructure, interrupt port operations, and disrupt cargo movements across these connected networks \citep{Asariotis2017_PortIndustrySurvey, verschuur2020port}. During the 2020 Atlantic hurricane season, nearly every port along the Gulf and South Atlantic coasts experienced TC-related operational shutdowns \citep{stacey2025port, li2023prediction}. When a single TC threatens several ports, anticipating which ports are likely to be affected and where disruptions may last longest can help authorities prioritize preparedness and recovery efforts.

As a TC approaches, ports and vessel operators take precautionary measures such as suspending port operations and rerouting vessels \citep{li2022ships, polydoropoulou2025composite}. Current hurricane preparedness protocols use a series of port conditions, including Whiskey, X-ray, Yankee, and Zulu, based on the expected arrival of gale-force winds \citep{USCG2024_PortCondition}. These conditions guide restrictions on vessel movements. After the winds subside, the Captain of the Port (COTP), in coordination with the Port Heavy Weather Advisory Group (PHWAG), assesses port facilities, navigational aids, and operational readiness before authorizing reopening \citep{USCG2024_PortCondition}. Such precautions reduce immediate safety risks, but closures can also result in substantial economic losses \citep{baker2012dynamic}. Forecast uncertainty complicates these decisions. An overly cautious response may impose unnecessary costs, while underestimating the threat may leave a port inadequately prepared \citep{mollaoglu2023bibliometric}. Moreover, ports exposed to the same TC can experience different operational impacts depending on the cyclone's intensity and landfall location as well as the characteristics of each port \citep{liu2026assessing, sturgis2014port}. A forecast track alone therefore provides limited information about how long operations at individual ports may be disrupted. Estimates of the relative impacts across ports could help authorities prioritize inspections and allocate recovery resources \citep{lin2006integrated, pu2025recent, wei2024supporting}. They could also help vessel operators evaluate rerouting options before the cyclone arrives \citep{edwards2023determining}.

Existing studies have used empirical data, including Automatic Identification System (AIS) records and port throughput data, to quantify port impacts and recovery under extreme weather events \citep{farhadi2016use, touzinsky2018using, trepte2014initial, verschuur2020port, kuai2026us}. Several studies use operational records to estimate disruption duration and construct resilience curves for affected ports \citep{touzinsky2018using, verschuur2020port}. These analyses establish how port activity declined and recovered during past events. For preparedness planning, the challenge is to estimate the potential impacts of an impending TC while its eventual path and the resulting disruptions are still unknown \citep{zhou2021analyzing}. To estimate port performance under such conditions, researchers have adopted qualitative methods such as the fuzzy analytical hierarchy process \citep{tahri2017using}, as well as quantitative approaches including simulation \citep{dhanak2021resilience, folkman2021port} and Bayesian networks \citep{hossain2020modeling, wang2023assessment}. These approaches have advanced the assessment of port performance under extreme weather, although their estimates can be sensitive to model assumptions and expert judgment.

The availability of vessel AIS data and advances in data-driven algorithms have expanded the use of artificial intelligence in port operations \citep{filom2022applications}. Many studies have employed machine learning (ML) and deep learning (DL) to predict routine operational outcomes, such as vessel arrival time \citep{el2023deep, park2021vessel} and port throughput \citep{li2019review}. Recent work has also examined how well AI models interpret cyclone forecast products for port preparedness \citep{kuai2026cyportqa, kuai2025knowledge,wu2026knowledge}. Estimating the operational impacts of an impending TC presents two challenges. First, TC-related disruptions are infrequent at individual ports, limiting the historical observations available for model training. Second, differences in impact duration across ports depend on interactions between port and cyclone characteristics, and uncertainty in the forecast landfall location makes these differences difficult to estimate before landfall.

This study examines port impact ranking in the Gulf of Mexico using data from five years between 2017 and 2022. We exclude 2020 because pandemic-related disruptions complicate the attribution of changes in transportation activity to TCs \citep{choi2021risk, kong2024case}. We use AIS vessel counts to identify the number of impact days at each port during each historical TC. We then combine these observations with port characteristics from the Port Performance Freight Statistics Program \citep{stacey2025port} and cyclone records from the National Hurricane Center \citep{NHC2023a_Graphics, NHC2023b_Glossary}. The resulting dataset links each observed port--cyclone interaction to its impact duration and to the characteristics of both the port and the cyclone.

Building on earlier work on port impact ranking \citep{li2023prediction}, we formulate the assessment of relative port impacts as a recommendation problem. Following the user--item formulation of recommendation systems \citep{ricci2021recommender}, each TC is treated as a user, each port as an item, and the observed number of impact days as the interaction signal. This formulation allows the model to draw on observations from ports and cyclones with similar characteristics when historical observations for a particular combination are limited. To account for forecast uncertainty, we augment the historical data by sampling plausible landfall locations along the coastline within the forecast uncertainty cone while retaining the observed number of impact days for each port--cyclone pair. For each landfall scenario, the two-stage recommendation framework first retrieves potentially affected ports and then ranks them by their estimated number of impact days. Within the ranking stage, we compare three representations of port--cyclone interactions: pairwise interactions in a factorization machine (FM), tree-based feature combinations in gradient boosted decision trees with linear regression (GBDT+LR), and explicit feature crosses combined with learned embeddings in Wide \& Deep Learning (WDL).

This study makes three contributions. First, we construct a dataset that links AIS-derived impact days with port and cyclone characteristics, providing an empirical basis for comparing the operational impacts of the same TC across ports. Second, we extend earlier FM-based work by examining how different representations of port--cyclone interactions can be used to rank potentially affected ports within a common two-stage recommendation framework when historical observations are sparse. Third, using Hurricanes Nicole and Ian as out-of-year test cases, we evaluate how well the framework identifies affected ports and how closely the estimated rankings agree with the observed port impact rankings across plausible landfall locations. This evaluation addresses the practical question of which ports should receive priority for post-cyclone inspections and recovery efforts.

\section{Data Collection and Processing}

In this study, our goal is to rank ports by their expected number of impact days under an impending tropical cyclone. An impact day is a day on which port activity falls below its normal operating range during a tropical cyclone. A port with a greater number of impact days is ranked as more severely affected. To estimate these relative impacts, the recommendation algorithm draws on historical interactions involving ports and cyclones with similar characteristics \citep{ricci2021recommender}. We treat each tropical cyclone as a user, each port as an item, and the observed number of impact days as the interaction signal \citep{li2023prediction}. This section describes the port and cyclone characteristics used in the analysis and explains how the number of impact days is estimated for each port--cyclone pair.

\subsection{Port Characteristics Data}

The Gulf Coast, with its abundant oil and gas reserves and strategic shipping routes, serves as a hub for the petrochemical industry and related activities in the United States. The region is frequently affected by tropical cyclones that disrupt port operations \citep{kuai2026us}. During Hurricane Laura in 2020, for example, several ports along the Gulf Coast partially or completely suspended operations. This study focuses on 37 ports in the study region. Information on port characteristics was manually collected and processed from the \textit{Port Performance Freight Statistics Program} \citep{stacey2025port}. Table~\ref{tab:port_characteristics} summarizes the location, governance, cargo activity, and harbor conditions of the study ports. Given the relative stability of these characteristics and the limited availability of annual observations, the same set of port attributes is used for the five study years.

\begin{table}[htbp]
\centering
\caption{Descriptive statistics of port infrastructure data}
\label{tab:port_characteristics}
\scriptsize
\renewcommand{\arraystretch}{1.1}
\setlength{\tabcolsep}{3pt}

\begin{tabularx}{\linewidth}{
    *{4}{>{\centering\arraybackslash}X}|
    *{4}{>{\centering\arraybackslash}X}
}
\toprule
\multicolumn{2}{l}{\textbf{Variable}}
& \textbf{Num.} & \textbf{Prop. (\%)}
& \multicolumn{2}{l}{\textbf{Variable}}
& \textbf{Num.} & \textbf{Prop. (\%)} \\
\midrule

\multicolumn{4}{l|}{\textbf{\textit{State}}}
& \multicolumn{4}{l}{\textbf{\textit{Dominant vessel type}}} \\

\multicolumn{2}{l}{Alabama (AL)} & 2 & 5.4
& \multicolumn{2}{l}{Tanker} & 15 & 40.5 \\

\multicolumn{2}{l}{Florida (FL)} & 8 & 21.6
& \multicolumn{2}{l}{Container} & 7 & 18.9 \\

\multicolumn{2}{l}{Louisiana (LA)} & 10 & 27.0
& \multicolumn{2}{l}{Dry bulk} & 13 & 35.1 \\

\multicolumn{2}{l}{Mississippi (MS)} & 6 & 16.2
& \multicolumn{2}{l}{Roll-on/roll-off} & 2 & 5.4 \\

\multicolumn{2}{l}{Texas (TX)} & 11 & 29.7
& \multicolumn{4}{l}{\textbf{\textit{Harbor type}}} \\

\multicolumn{4}{l|}{\textbf{\textit{Port governance}}}
& \multicolumn{2}{l}{River natural} & 20 & 54.1 \\

\multicolumn{2}{l}{Other} & 3 & 8.1
& \multicolumn{2}{l}{Coastal natural} & 11 & 29.7 \\

\multicolumn{2}{l}{Municipal} & 3 & 8.1
& \multicolumn{2}{l}{Lake or canal} & 4 & 10.8 \\

\multicolumn{2}{l}{Special district} & 13 & 35.1
& \multicolumn{2}{l}{Sea port} & 2 & 5.4 \\

% Verify whether the governance category in the source data is Country or County.
\multicolumn{2}{l}{Country} & 13 & 35.1
& \multicolumn{4}{l}{\textbf{\textit{Harbor size}}} \\

\multicolumn{2}{l}{State} & 5 & 13.5
& \multicolumn{2}{l}{Very small} & 7 & 18.9 \\

\multicolumn{4}{l|}{\textbf{\textit{Container port}}}
& \multicolumn{2}{l}{Small} & 16 & 43.2 \\

\multicolumn{2}{l}{Yes} & 12 & 32.4
& \multicolumn{2}{l}{Medium} & 10 & 27.0 \\

\multicolumn{2}{l}{No} & 25 & 67.6
& \multicolumn{2}{l}{Large} & 4 & 10.8 \\

\multicolumn{4}{l|}{\textbf{\textit{Annual average tonnage (million tons)}}}
& \multicolumn{4}{l}{\textbf{\textit{Shelter condition}}} \\

Mean & Std. & Min & Max
& \multicolumn{2}{l}{Average} & 7 & 18.9 \\

35.09 & 61.09 & 0.99 & 275.94
& \multicolumn{2}{l}{Good} & 21 & 56.8 \\

& & & &
\multicolumn{2}{l}{Excellent} & 9 & 24.3 \\

\bottomrule
\end{tabularx}

\medskip
\begin{minipage}{\linewidth}
\scriptsize
\textit{Note:} Num. denotes the number of ports, and Prop. denotes
the percentage of the 37 study ports. Percentages are rounded to
one decimal place. Mean, standard deviation (Std.), minimum (Min),
and maximum (Max) are reported for annual average tonnage.
\end{minipage}
\end{table}

\subsection{Tropical Cyclone Data}

We collect data on named tropical cyclones in the study region between 2017 and 2022, excluding 2020. Table~\ref{tab:cyclone_features} summarizes 31 cyclones across the five study years. The \textit{National Hurricane Center (NHC)} provides trajectory records and meteorological parameters for each cyclone, including central pressure and wind speed \citep{cangialosi2020recent}. Given the relationship between wind speed and cyclone destructiveness, we focus on wind-related attributes \citep{emanuel2005increasing,kuai2026us}. The trajectory records include the dates and times of formation, landfall, and dissipation, as well as the coordinates of key locations along each track. We use these spatial and temporal features to derive variables for each port--cyclone pair, including the distance from the port to the landfall location. For cyclones that did not make landfall, we use the point on the track closest to the coastline as the reference location and calculate the distance from each port to that point. For cyclones that made multiple landfalls, we use the first landfall location.

\begin{table}[htbp]
\centering
\caption{Descriptive statistics of tropical cyclone data}
\label{tab:cyclone_features}
\scriptsize
\renewcommand{\arraystretch}{1.1}
\setlength{\tabcolsep}{3pt}

\begin{tabularx}{\linewidth}{
    *{4}{>{\centering\arraybackslash}X}|
    *{4}{>{\centering\arraybackslash}X}
}
\toprule
\multicolumn{2}{l}{\textbf{Variable}}
& \textbf{Num.} & \textbf{Prop. (\%)}
& \multicolumn{2}{l}{\textbf{Variable}}
& \textbf{Num.} & \textbf{Prop. (\%)} \\
\midrule

\multicolumn{4}{l|}{\textbf{\textit{Year}}}
& \multicolumn{4}{l}{\textbf{\textit{Tropical cyclone intensity}}} \\

\multicolumn{2}{l}{2017} & 8 & 25.8
& \multicolumn{2}{l}{TS} & 12 & 38.7 \\

\multicolumn{2}{l}{2018} & 5 & 16.1
& \multicolumn{2}{l}{H1} & 8 & 25.8 \\

\multicolumn{2}{l}{2019} & 6 & 19.4
& \multicolumn{2}{l}{H2} & 2 & 6.5 \\

\multicolumn{2}{l}{2021} & 7 & 22.6
& \multicolumn{2}{l}{H3} & 3 & 9.7 \\

\multicolumn{2}{l}{2022} & 5 & 16.1
& \multicolumn{2}{l}{H4} & 3 & 9.7 \\

& & & &
\multicolumn{2}{l}{H5} & 3 & 9.7 \\

\multicolumn{4}{l|}{\textbf{\textit{Lowest central pressure (millibars)}}}
& \multicolumn{4}{l}{\textbf{\textit{Maximum wind speed (kts)}}} \\

Mean & Std. & Min & Max
& Mean & Std. & Min & Max \\

976.3 & 26.59 & 914 & 1003
& 77.6 & 34.2 & 35 & 155 \\

\multicolumn{4}{l|}{\textbf{\textit{Wind speed at landfall (kts)}}}
& \multicolumn{4}{l}{\textbf{\textit{Duration from formation to landfall (days)}}} \\

Mean & Std. & Min & Max
& Mean & Std. & Min & Max \\

65.8 & 28.9 & 35 & 140
& 4.4 & 3.3 & 0.71 & 15.5 \\

\multicolumn{4}{l|}{\textbf{\textit{Duration from formation to dissipation (days)}}}
& \multicolumn{4}{l}{\textbf{\textit{Distance from port to formation location (km)}}} \\

Mean & Std. & Min & Max
& Mean & Std. & Min & Max \\

6.7 & 4.3 & 2 & 19
& 1845.8 & 1570.7 & 42.2 & 7465.0 \\

\multicolumn{4}{l|}{\textbf{\textit{Distance from port to maximum wind location (km)}}}
& \multicolumn{4}{l}{\textbf{\textit{Distance from port to landfall location (km)}}} \\

Mean & Std. & Min & Max
& Mean & Std. & Min & Max \\

1131.0 & 823.6 & 8.7 & 3932.2
& 867.7 & 489.0 & 0.2 & 2272.3 \\

\bottomrule
\end{tabularx}

\medskip
\begin{minipage}{\linewidth}
\scriptsize
\textit{Note:} TS denotes a tropical storm, and H1--H5 denote
Categories 1--5 on the Saffir--Simpson Hurricane Wind Scale.
Wind speed is reported in knots (kts), where 1 knot = 1.852 km/h.
Num. denotes the number of cyclones, and Prop. denotes the
percentage of the 31 cyclones summarized in the table.
The study years are 2017, 2018, 2019, 2021, and 2022.
\end{minipage}
\end{table}

\subsection{AIS Data}

To determine the number of impact days for each port--cyclone pair, we use the number of commercial vessels within each port area as an indicator of port performance \citep{touzinsky2018using}. Automatic Identification System (AIS) data record vessel locations and operational characteristics and are widely used for maritime applications such as collision avoidance, vessel assignment, and port management \citep{yang2019big}. The records include timestamps, coordinates, vessel types, speeds, and headings at approximately 30-second intervals. A single year of AIS data can exceed 300~GB. As in other analyses of transportation activity records, preprocessing reduces the volume of data and prepares the observations for analysis \citep{li2019risk}. Following \citet{verschuur2020port}, we apply the following steps:

\begin{itemize}[leftmargin=1.5em, itemsep=0pt, parsep=0pt, topsep=2pt]
    \item We retain cargo vessels, tankers, and container vessels and
    exclude vessel types outside the scope of the analysis, including
    fishing vessels, tugs, and towing vessels.
    \item We define port boundaries using coordinate ranges and count
    the active vessels within each boundary.
    \item We aggregate hourly vessel counts into a daily time series
    for each port.
\end{itemize}

The resulting time series are used to estimate the number of impact days at each port. Previous studies have applied threshold methods \citep{verschuur2020port} or Bayesian change-point detection \citep{touzinsky2018using} to identify impact periods. To account for recurring seasonal variation in port activity, we use a modular time-series regression model \citep{taylor2018forecasting}. The model decomposes the vessel count series into trend and seasonal components. We adopt an additive formulation so that the expected daily vessel count reflects both the underlying trend and the seasonal pattern at each port. For each port, we fit the additive regression model using one year of vessel count data. A day is classified as an impact day when the observed vessel count falls below the lower bound of the model's predefined interval and the date coincides with a tropical cyclone event.

\begin{figure}[htbp]
    \centering
    \includegraphics[width=0.83\textwidth]{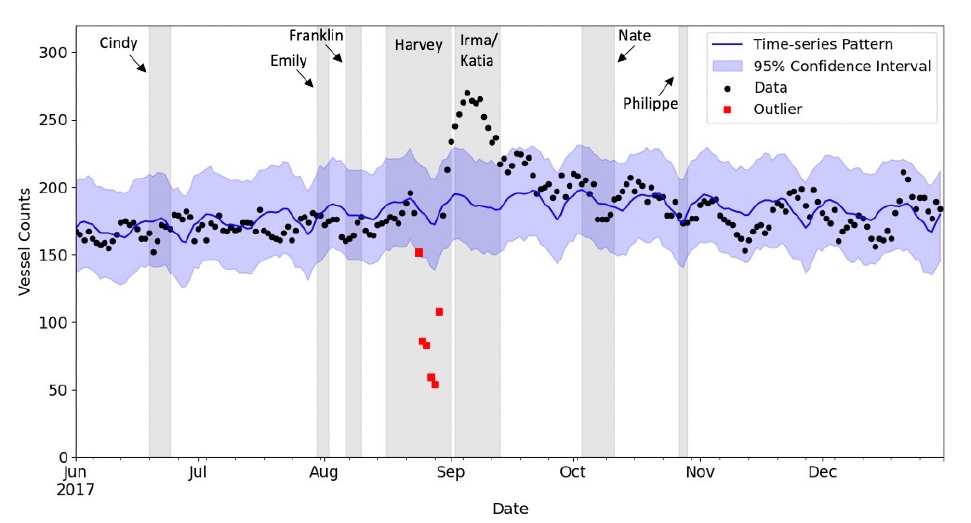}
    \caption{Identifying cyclone-related impact days from vessel counts at Port Houston in 2017.}
    \label{fig:additive_regression}
\end{figure}

Figure~\ref{fig:additive_regression} illustrates the fitted time-series pattern and the 95\% interval for Port Houston in 2017. During that hurricane season, only Hurricane Harvey was associated with vessel counts low enough to meet the impact-day criterion, resulting in six impact days. Days with vessel counts above the upper bound are excluded from the impact-day count. Such increases may reflect a temporary rise in activity as ports handle cargo delayed by an earlier disruption \citep{verschuur2020port}. Across the port--cyclone pairs analyzed during 2017--2022, excluding 2020, the mean number of impact days was 0.29, with a standard deviation of 0.95. The observed values ranged from 0 to 7 days.

\section{Methodology}

Historical observations of port disruptions under tropical cyclones (TCs) are sparse. As described in Section~2, we treat each TC as a user, each port as an item, and the observed number of impact days as the interaction signal \citep{ricci2021recommender,li2023prediction}. This formulation allows the model to draw on observations from ports and cyclones with similar characteristics when estimating the impact of an impending TC.

\begin{figure}[htbp]
    \centering
    \includegraphics[width=0.9\textwidth]{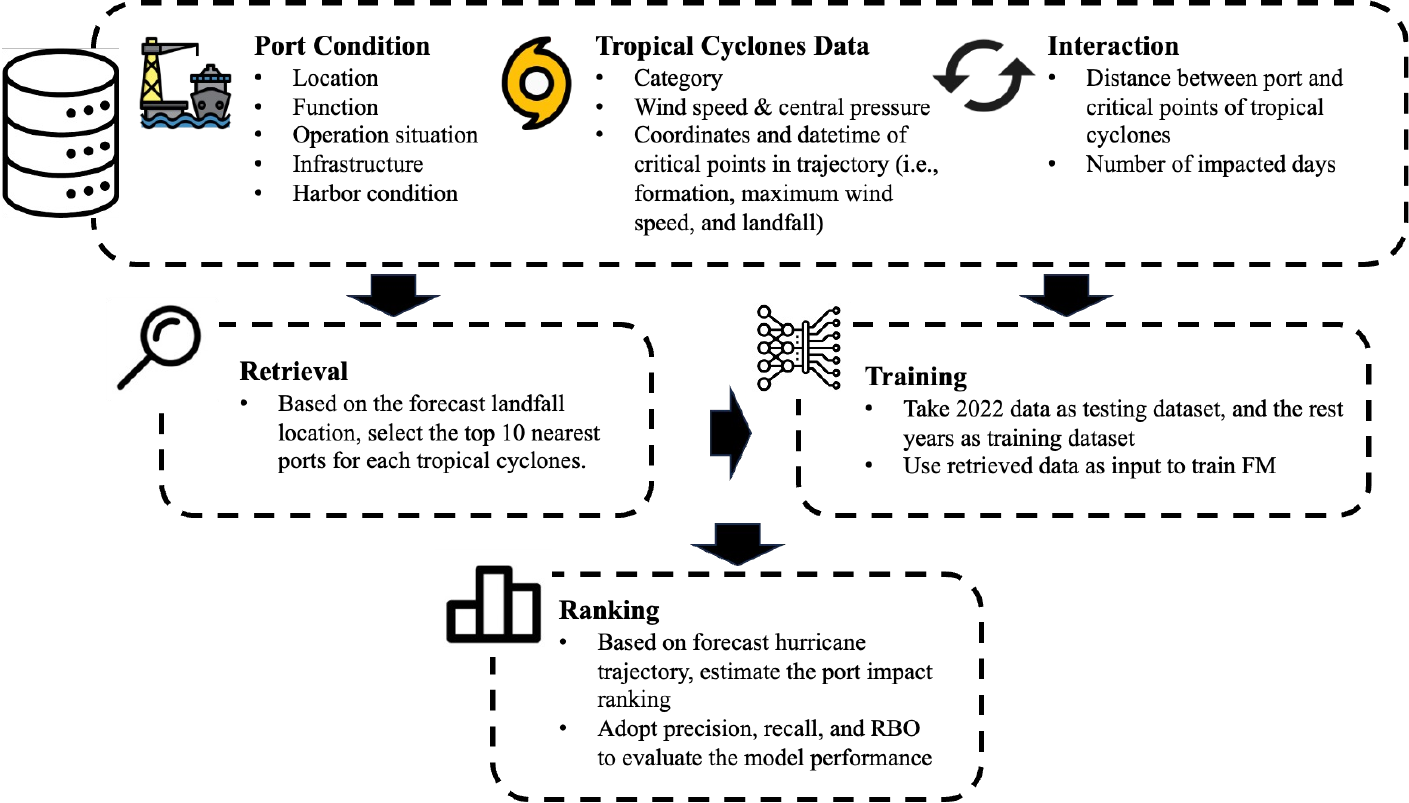}
    \caption{Two-stage recommendation framework for port impact ranking.}
    \label{fig:framework}
\end{figure}

Using historical data on 37 ports and 31 TCs, we apply a two-stage recommendation framework to identify potentially affected ports and rank them by their expected number of impact days \citep{li2023prediction}. For each plausible landfall location, we first retrieve the ten nearest ports. We then estimate the number of impact days for each retrieved port and rank the ports in descending order of their estimates. All three ranking models use the same candidate ports within a given landfall scenario (Figure~\ref{fig:framework}).

Before a TC makes landfall, uncertainty in its track affects which ports may be exposed \citep{cangialosi2020recent,NHC2023a_Graphics}. To represent uncertainty in landfall location in the historical data, we define a sampling region with a radius of 120~km centered on the recorded landfall point \citep{li2023prediction}. For each TC, we randomly sample 20 plausible landfall locations uniformly along the portion of the coastline within this region. Each sampled location defines one landfall scenario. We recalculate the distance from each of the 37 ports to the sampled point and use these distances to retrieve the ten nearest ports.

\begin{figure}[htbp]
    \centering
    \includegraphics[width=0.85\textwidth]{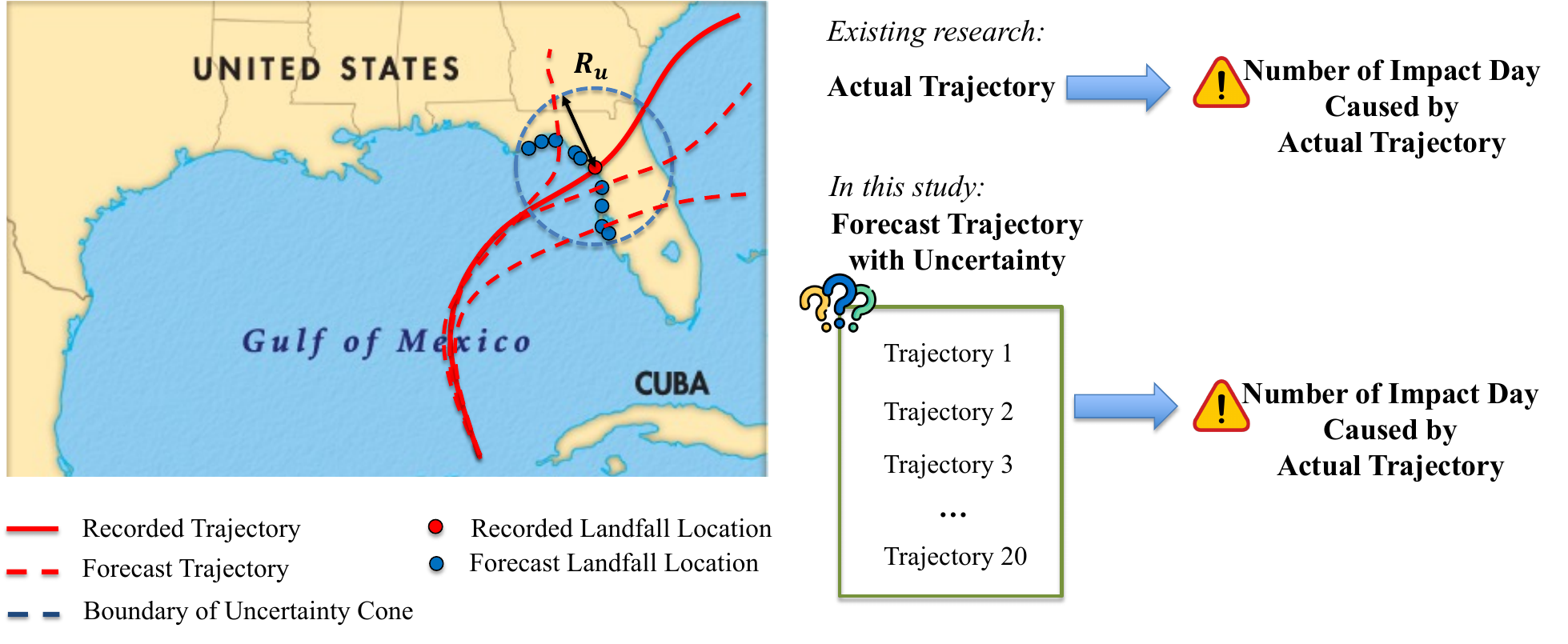}
    \caption{Data augmentation using plausible landfall locations sampled along the coastline within 120~km of the recorded landfall point.}
    \label{fig:augmentation}
\end{figure}

This data augmentation varies the landfall location while retaining the observed impacts of each historical event \citep{rebuffi2021data}. As the sampled location changes, the distances to the ports change, and a different set of ports may be retrieved. The observed number of impact days for each port--cyclone pair remains fixed across scenarios because it is derived from AIS vessel counts during the historical event. This allows the rankings obtained under different plausible landfall locations to be evaluated against the port impacts observed during the same TC (Figure~\ref{fig:augmentation}).

Across the 31 TCs, the 20 sampled locations per cyclone yield 620 landfall scenarios. Before retrieval, the augmented dataset contains $37 \times 31 \times 20 = 22{,}940$ port--cyclone--scenario records. Selecting ten ports per scenario yields $10 \times 31 \times 20 = 6{,}200$ candidate records across the full dataset. We use data from 2017, 2018, 2019, and 2021 for model development and Hurricanes Nicole and Ian in 2022 as the two test cases. All scenarios associated with a given cyclone are kept together when partitioning the data.

\subsection{Recommendation Algorithm}
\label{rec_alg}

The three ranking models differ in how they represent interactions between port and cyclone characteristics. Feature crossing allows combinations of attributes to contribute to the estimated number of impact days \citep{liu2023stationary}. A factorization machine (FM) represents pairwise feature interactions through latent vectors \citep{rendle2010factorization}. Gradient boosted decision trees combined with linear regression (GBDT+LR) use combinations of conditions along tree paths to generate features for regression. Wide \& Deep Learning (WDL) jointly learns from explicit feature crosses and dense embeddings \citep{cheng2016wide}. Each model estimates impact days for the retrieved ports, allowing their rankings to be compared within the same landfall scenario.

\subsubsection{Factorization Machine}

A factorization machine captures second-order feature interactions by factorizing their weights, making it suitable for sparse, high-dimensional inputs \citep{rendle2010factorization}. In our implementation, continuous port and cyclone attributes are discretized before one-hot encoding. The feature-processing details are provided in Appendix~\ref{app:implementation}.

The FM consists of a linear component and a pairwise interaction component. Each input feature $x_i$ is associated with a $k$-dimensional latent vector $\mathbf{v}_i$, where $k$ is a hyperparameter. The interaction weight for features $x_i$ and $x_j$ is given by their inner product,
$\langle \mathbf{v}_i,\mathbf{v}_j\rangle =
\sum_{f=1}^{k}v_{i,f}v_{j,f}$.
The same latent vector $\mathbf{v}_i$ is used in every interaction involving feature $x_i$. Thus, observations of other feature combinations contribute to the estimated interaction weight even when a particular pair is rarely observed. For $n$ input features, the interaction component uses $nk$ latent parameters in place of a separate weight for each of the $n(n-1)/2$ feature pairs. The estimated number of impact days is

\begin{equation}
\hat{y}(\mathbf{x}) =
\underbrace{\left(
w_0 + \sum_{i=1}^{n} w_i x_i
\right)}_{\text{linear component}}
+
\underbrace{\left(
\sum_{i=1}^{n-1}\sum_{j=i+1}^{n}
\langle \mathbf{v}_i,\mathbf{v}_j\rangle x_i x_j
\right)}_{\text{pairwise interactions}},
\label{eq:fm1}
\end{equation}

\noindent
where $w_0$ is the bias term and $w_i$ is the linear weight for feature $x_i$. Rewriting Eq.~(\ref{eq:fm1}) as Eq.~(\ref{eq:fm2}) reduces the computational complexity of evaluating the interaction terms from $O(kn^2)$ to $O(kn)$ \citep{rendle2010factorization}.

\begin{equation}
\hat{y}(\mathbf{x}) =
w_0 + \sum_{i=1}^{n}w_ix_i
+
\frac{1}{2}\sum_{f=1}^{k}
\left[
\left(\sum_{i=1}^{n}v_{i,f}x_i\right)^2
-
\sum_{i=1}^{n}v_{i,f}^{2}x_i^{2}
\right].
\label{eq:fm2}
\end{equation}

We estimate the model parameters using stochastic gradient descent (SGD) with squared-error loss.

\subsubsection{Gradient Boosted Decision Trees with Linear Regression}

\begin{figure}[htbp]
    \centering
    \includegraphics[width=1\linewidth]{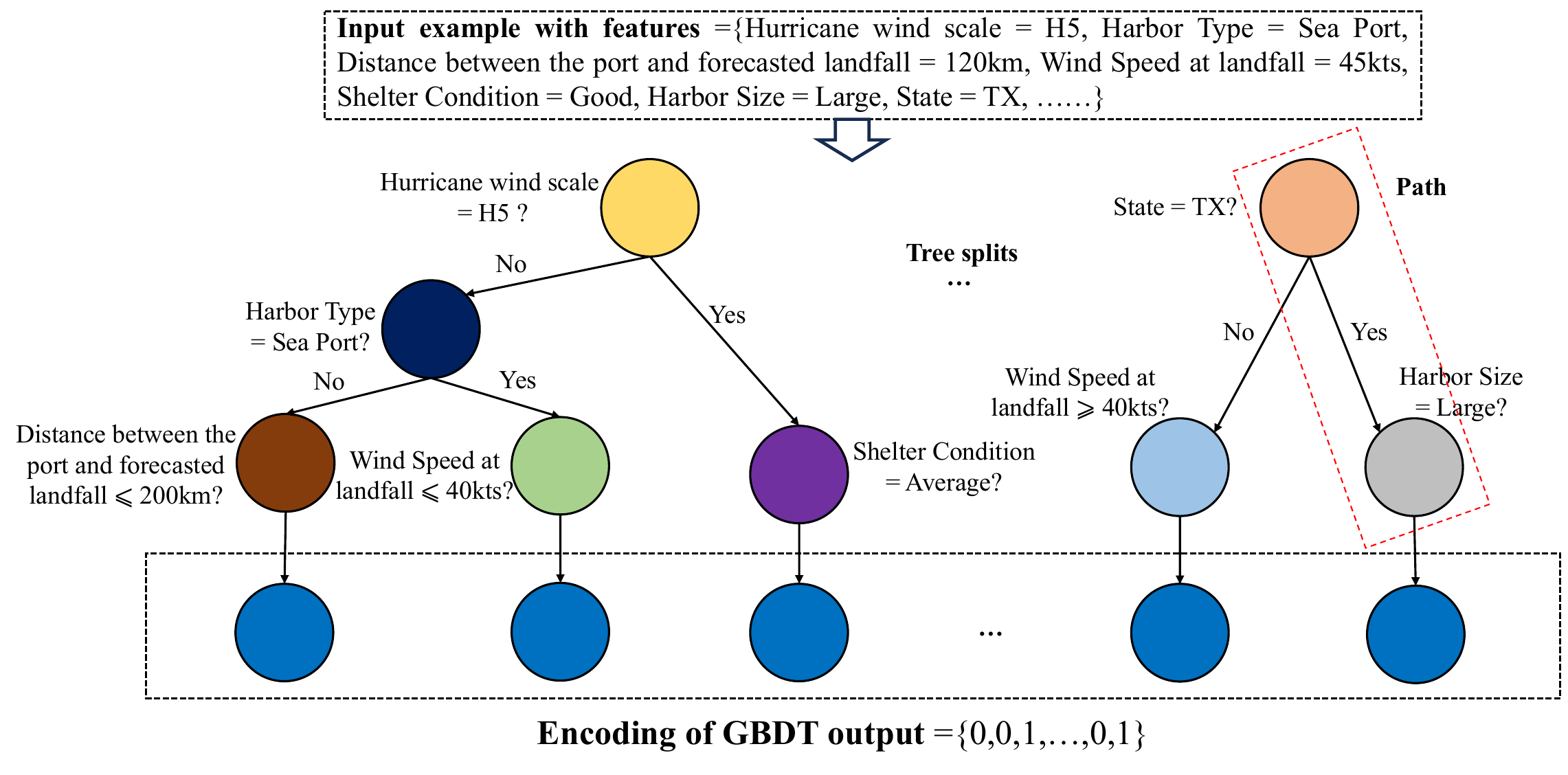}
    \caption{Tree-based encoding of port and cyclone feature combinations.}
    \label{fig:gbdt_embedding}
\end{figure}

Gradient boosted decision trees can capture nonlinear relationships among input features \citep{friedman2001greedy,kong2022bridge,li2023applying,li2023exploring,kong2024investigating, li2025leveraging}. In our GBDT+LR model, the trees generate features from port and cyclone characteristics, and linear regression uses these features to estimate impact days. The trees are fitted iteratively, with successive trees using the negative gradients of the boosting loss to update the model \citep{friedman2001greedy}. Each path from the root of a tree to a leaf represents a combination of conditions on the input attributes. For example, one path combines three conditions: the hurricane wind scale is below H5, the harbor type is not a seaport, and the port is less than 200~km from the sampled landfall location (Figure~\ref{fig:gbdt_embedding}). A port--cyclone pair that satisfies the conditions along this path reaches the corresponding leaf.

\begin{figure}[htbp]
    \centering
    \includegraphics[width=0.75\linewidth]{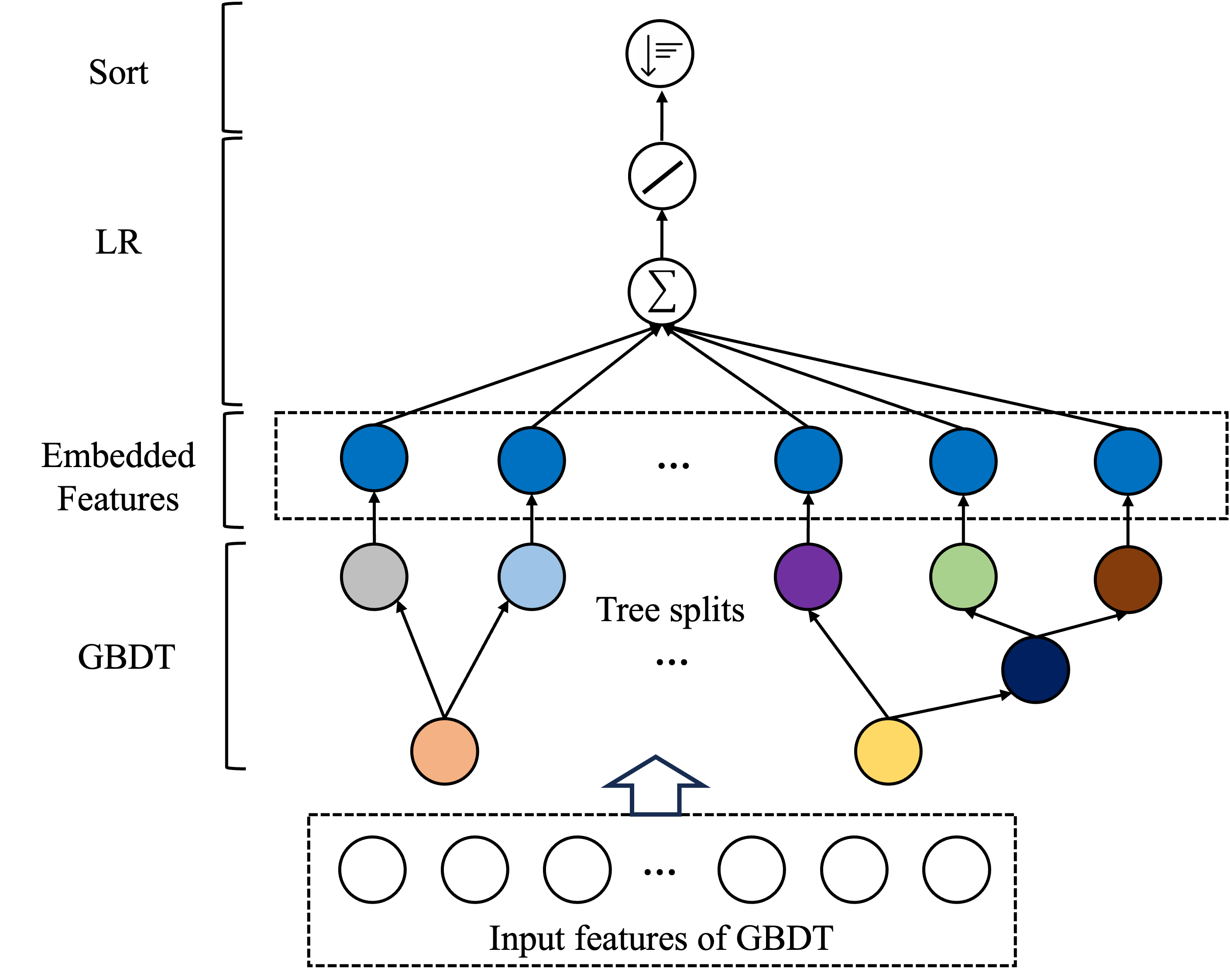}
    \caption{Port impact estimation using GBDT leaf encodings and linear regression.}
    \label{fig:gbdt_framework}
\end{figure}

For each training example, we record the index of the leaf reached in each selected tree and apply one-hot encoding to the leaf indices. The resulting binary vector has one active entry per selected tree. Each active entry identifies a combination of feature conditions satisfied by that example. These vectors serve as inputs to a linear regression model, with the observed number of impact days as the target. For each new landfall scenario, the fitted trees and the same leaf encoder are applied to the retrieved ports before estimating their impact days (Algorithm~\ref{alg:gbdt_lr} and Figure~\ref{fig:gbdt_framework}).

\begin{algorithm}[htbp]
\caption{GBDT with linear regression for port impact ranking}
\label{alg:gbdt_lr}
\begin{algorithmic}[1]
\Require Training examples $\{(\mathbf{x}_i,y_i)\}_{i=1}^{N}$,
number of boosting iterations $T$, maximum tree depth $D$,
and learning rate $\eta$
\Ensure Estimated number of impact days for each retrieved port
\State Fit a gradient boosting classifier to the training examples
using $T$ boosting iterations, maximum depth $D$, and learning rate $\eta$
\State Select the trees used for leaf encoding as described in
Appendix~\ref{app:implementation}
\For{$i \leftarrow 1$ to $N$}
    \State Record the leaf index reached by $\mathbf{x}_i$ in each
    selected tree as the vector $\boldsymbol{\ell}_i$
\EndFor
\State Fit a one-hot encoder $E$ to the training leaf-index vectors
\State Set $\mathbf{z}_i = E(\boldsymbol{\ell}_i)$ for each training example
\State Estimate $\mathbf{w}$ and $b$ by least squares using
$\{(\mathbf{z}_i,y_i)\}_{i=1}^{N}$
\For{each retrieved port in a landfall scenario}
    \State Obtain its leaf-index vector $\boldsymbol{\ell}$ from the fitted trees
    \State Estimate impact days as
    $\hat{y} = \mathbf{w}^{\top}E(\boldsymbol{\ell}) + b$
\EndFor
\end{algorithmic}
\end{algorithm}

\newpage

\subsubsection{Wide \& Deep Learning}

In GBDT+LR, the trees are fitted before the linear regression, so the leaf encodings remain fixed when the regression weights are estimated. Wide \& Deep Learning provides a third way to represent port--cyclone interactions by training a wide linear component and a deep neural network jointly \citep{cheng2016wide}. The wide component uses the original features and explicit feature crosses to capture associations between recurring feature combinations and observed impact days. The deep component maps categorical port and cyclone attributes to dense embedding vectors. These embeddings allow the model to draw on information from port--cyclone pairs with similar characteristics when historical observations for a particular combination are limited.

The embedding vectors are randomly initialized and updated through backpropagation during training. The concatenated embeddings form the input to the hidden layers of the deep component. The transformation from layer $l$ to layer $l+1$ is given by

\begin{equation}
a^{(l+1)} = f\left(W^{(l)}a^{(l)} + b^{(l)}\right),
\label{eq:3}
\end{equation}

\noindent
where $a^{(l)}$ is the activation vector at layer $l$, $W^{(l)}$ and $b^{(l)}$ are the weights and biases for the transformation, and $f(\cdot)$ is the activation function. Successive nonlinear transformations allow the deep component to learn interactions among the embedded features.

\begin{figure}[htbp]
    \centering
    \includegraphics[width=0.95\textwidth]{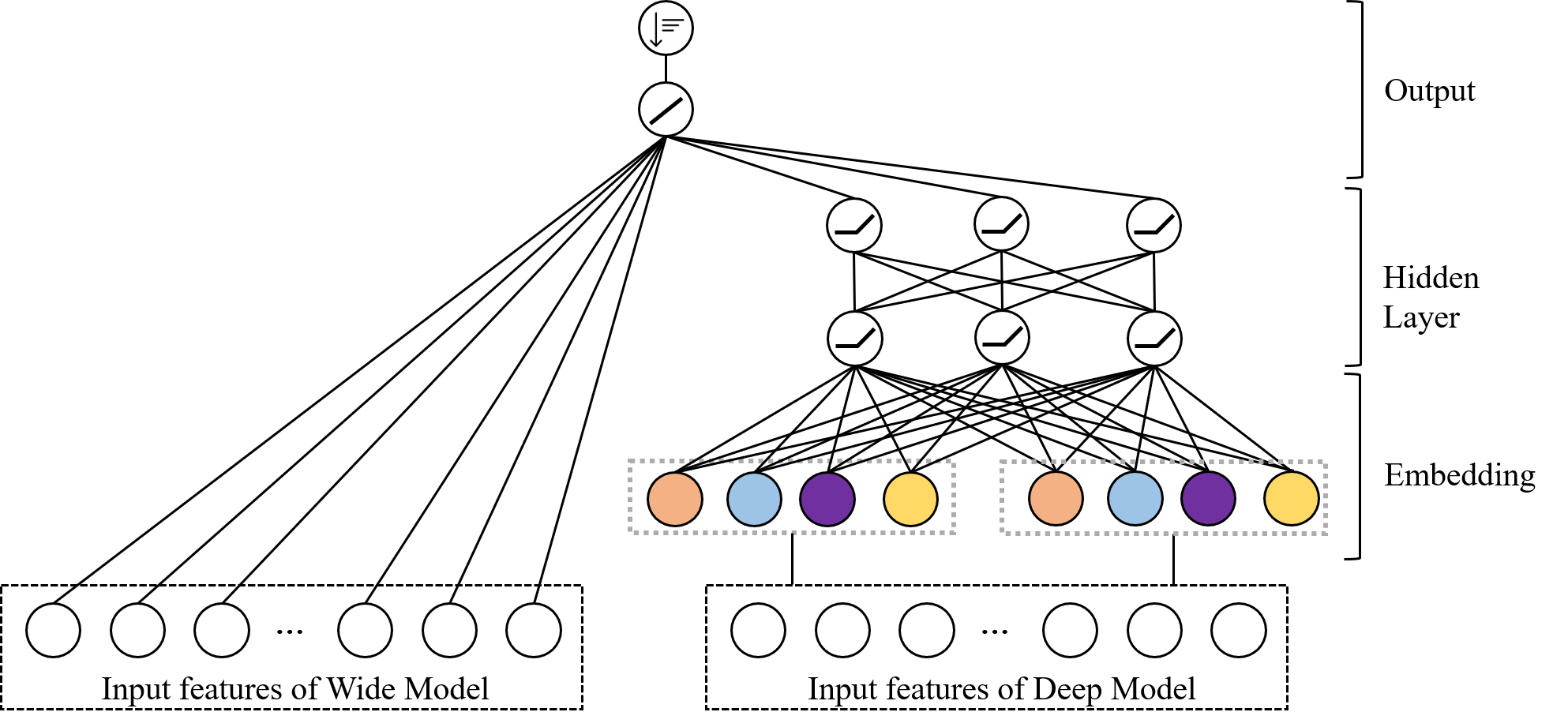}
    \caption{WDL architecture combining sparse feature crosses and dense embeddings.}
    \label{fig:wdl}
\end{figure}

To estimate impact days, the model combines the contribution of the original features and their explicit crosses with the output of the final hidden layer (Figure~\ref{fig:wdl}):

\begin{equation}
\hat{y}_{\mathrm{WDL}}(\mathbf{x})
=
w_{\mathrm{wide}}^{\top}
[\mathbf{x},\phi(\mathbf{x})]
+
w_{\mathrm{deep}}^{\top}a^{(l_f)}
+
b,
\label{eq:4}
\end{equation}

\noindent
where $\phi(\mathbf{x})$ denotes the explicit feature crosses and
$[\mathbf{x},\phi(\mathbf{x})]$ denotes their concatenation with the original features. The vectors $w_{\mathrm{wide}}$ and $w_{\mathrm{deep}}$ are the output weights for the wide component and the final deep-layer activation $a^{(l_f)}$, respectively, and $b$ is the output bias. During training, $L_2$ regularization penalizes large weights. As with FM and GBDT+LR, the estimated number of impact days serves as the ranking score for each retrieved port.

\section{Results and Discussion}

\subsection{Experimental Setup and Baseline Method}

We split the data by tropical cyclone, keeping all observations from the same TC, including its 20 landfall scenarios, in the same partition. Among the TCs from 2017, 2018, 2019, and 2021, approximately 20\% are reserved for validation and the remainder are used for training. Hurricanes Nicole and Ian from 2022 are held out for testing. For each test hurricane, FM, GBDT+LR, and WDL are evaluated on the same candidate ports within each landfall scenario. We implement the three models using PyFM, scikit-learn, and PyTorch, respectively. The model settings are summarized in Table~\ref{tab:hyperparams}, with implementation details provided in~\ref{app:implementation}.

\begin{table}[htbp]
\centering
\caption{Hyperparameter settings for the ranking models.}
\label{tab:hyperparams}
\renewcommand{\arraystretch}{1.05}
\setlength{\tabcolsep}{5pt}
\small
\begin{tabularx}{\linewidth}{@{}Xr@{\hspace{1.5em}}Xr@{}}
\toprule
\textbf{Parameter} & \textbf{Setting}
& \textbf{Parameter} & \textbf{Setting} \\
\midrule
\multicolumn{2}{@{}l}{\textbf{FM}}
& \multicolumn{2}{l}{\textbf{WDL}} \\

Number of latent factors ($k$) & 100
& Hidden-layer units & (256, 128, 64) \\

Initial learning rate & 0.001
& $L_2$ coefficient (wide component) & $10^{3}$ \\

\cmidrule(r){1-2}
\multicolumn{2}{@{}l}{\textbf{GBDT+LR}}
& Batch size & 256 \\

Number of trees & 120
& Training epochs & 100 \\

Maximum tree depth ($D$) & 30
& Validation split ratio & 0.2 \\

Learning rate ($\eta$) & 0.05
& Optimizer & Adam \\
\bottomrule
\end{tabularx}
\end{table}
\newpage

\begin{figure}[htbp]
\vspace{-6pt}
    \centering
    \includegraphics[width=0.65\textwidth]{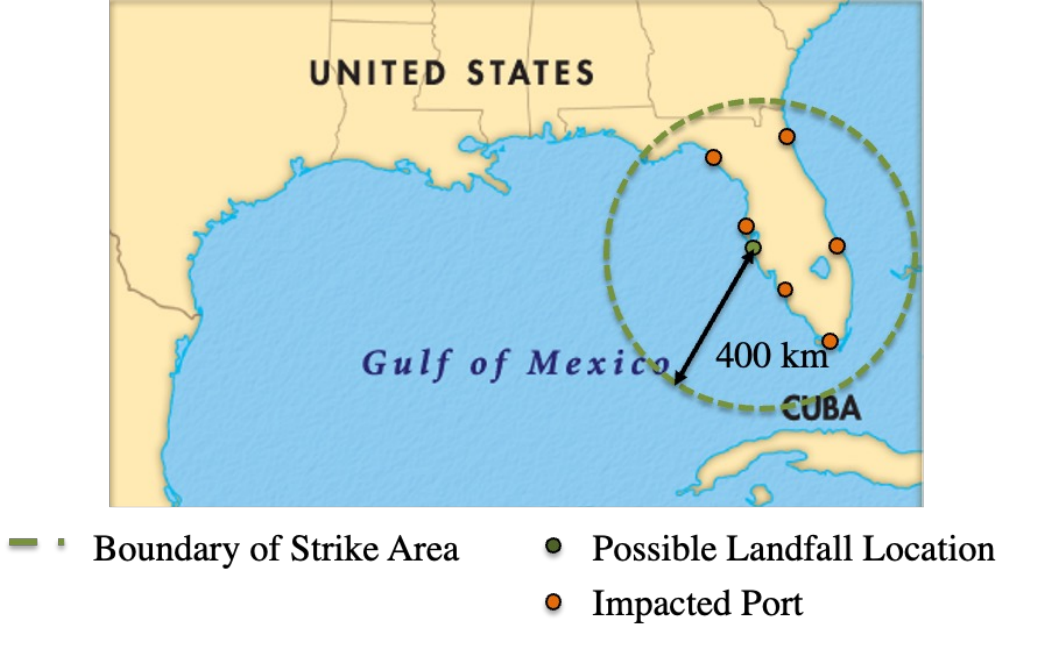}
    \caption{Distance-based screening and ranking of potentially affected ports.}
    \label{fig:distance}
\end{figure}

We use a distance-based method as the baseline for port impact ranking \citep{balakrishnan2022methodology,li2023prediction}. For each scenario, the baseline selects ports within 400~km of the sampled landfall location and ranks them from nearest to farthest (Figure~\ref{fig:distance}).

\vspace{-16pt}

\subsection{Performance Measures}

We evaluate retrieval using precision, recall, and the $F_2$ score. Let $U$ denote the set of ports with at least one observed impact day and $L$ the set of ports identified as potentially affected in a given landfall scenario. Precision ($P$) measures the proportion of selected ports that were affected, while recall ($R$) measures the proportion of affected ports that were selected:

\begin{equation}
P = \frac{|U \cap L|}{|L|},
\label{eq:port_precision}
\end{equation}

\begin{equation}
R = \frac{|U \cap L|}{|U|}.
\label{eq:port_recall}
\end{equation}

The recommendation framework retrieves ten ports in each scenario, while the number selected by the distance-based baseline varies with the sampled landfall location. Because overlooking an affected port is a concern for preparedness planning, we use the $F_{\beta}$ score with $\beta=2$ to give greater weight to recall:

\begin{equation}
F_{\beta}
=
\frac{(1+\beta^{2})PR}{\beta^{2}P+R}.
\label{eq:port_fbeta}
\end{equation}

For the ranking models, mean absolute error (MAE) and root mean square error (RMSE) measure the difference between the observed number of impact days, $y_i$, and the estimated number, $\hat{y}_i$, for the $N$ retrieved ports in each scenario:

\begin{equation}
\mathrm{MAE}
=
\frac{1}{N}\sum_{i=1}^{N}
\left|y_i-\hat{y}_i\right|,
\label{eq:port_mae}
\end{equation}

\begin{equation}
\mathrm{RMSE}
=
\sqrt{\frac{1}{N}\sum_{i=1}^{N}
\left(y_i-\hat{y}_i\right)^2}.
\label{eq:port_rmse}
\end{equation}

Both errors are expressed in days. The distance-based baseline provides only a ranking, so MAE and RMSE are reported for the three learned models.

To assess whether the ports with the longest observed disruptions appear near the top of the estimated ranking, we compare the overlap between the two lists at successive depths \citep{webber2010similarity}. Under the equal-weight setting ($p=1$) of the RBO implementation, the score is the average overlap (AO) across the depths shared by the two lists. Let $\mathcal{U}$ denote the observed ranking of affected ports by impact days and $\mathcal{L}$ the estimated ranking for the same scenario. At depth $d$, the sets $U_{1:d}$ and $L_{1:d}$ contain the first $d$ ports in the respective lists. The average overlap is

\begin{equation}
\mathrm{AO}@D
=
\frac{1}{D}\sum_{d=1}^{D}
\frac{\left|U_{1:d}\cap L_{1:d}\right|}{d},
\qquad
D=\min\!\left(|\mathcal{U}|,|\mathcal{L}|\right).
\label{eq:port_average_overlap}
\end{equation}

Each depth contributes equally to the average. Ports that appear near the top of both lists contribute to their overlap at multiple depths. The score ranges from 0 to 1, with higher values indicating closer agreement over the evaluated depths. For each test hurricane, we report the mean and standard deviation across its 20 sampled landfall locations.

\subsection{Comparison of Classification Performance}

\begin{figure}[htbp]
\vspace{-12pt}
  \centering
  \begin{subfigure}[t]{0.47\textwidth}
    \centering
    \includegraphics[width=\linewidth]{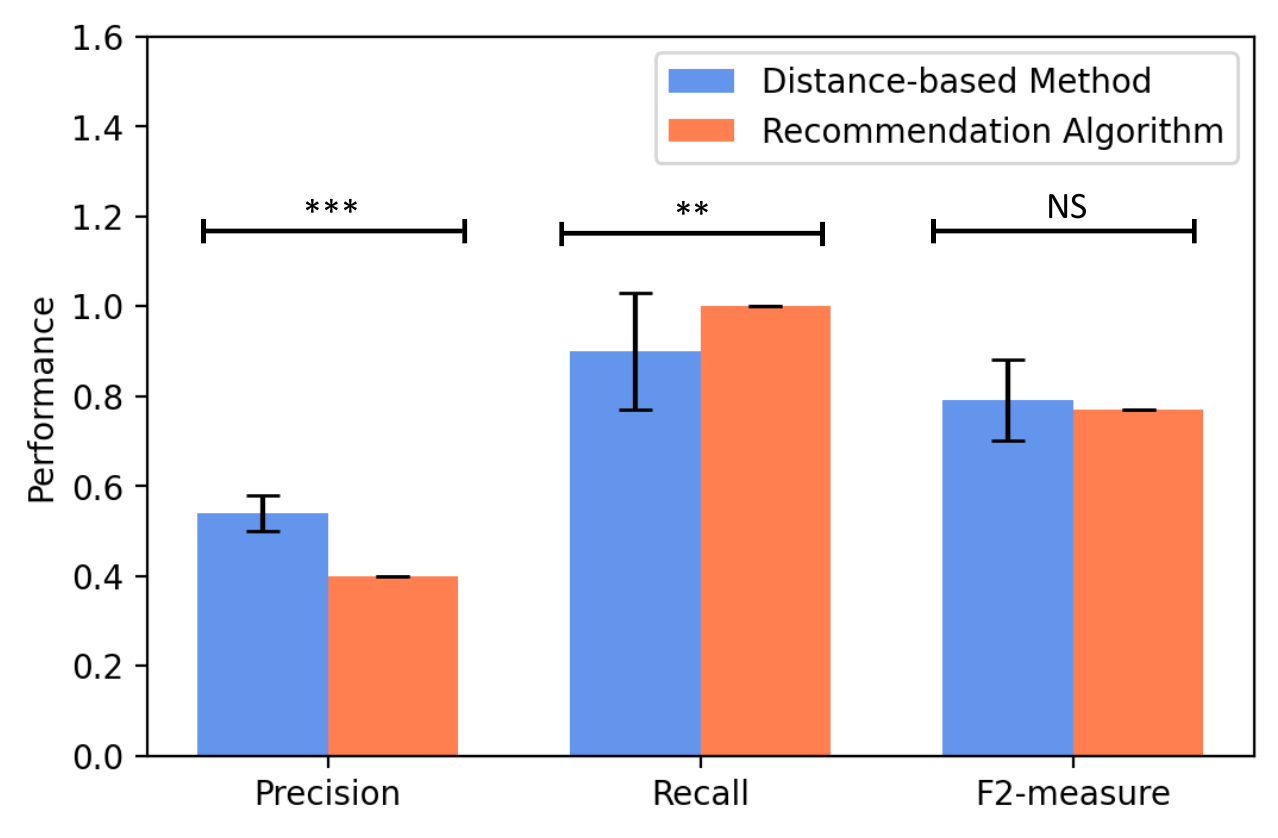}
    \caption{Hurricane Nicole (2022)}
  \end{subfigure}\hfill
  \begin{subfigure}[t]{0.47\textwidth}
    \centering
    \includegraphics[width=\linewidth]{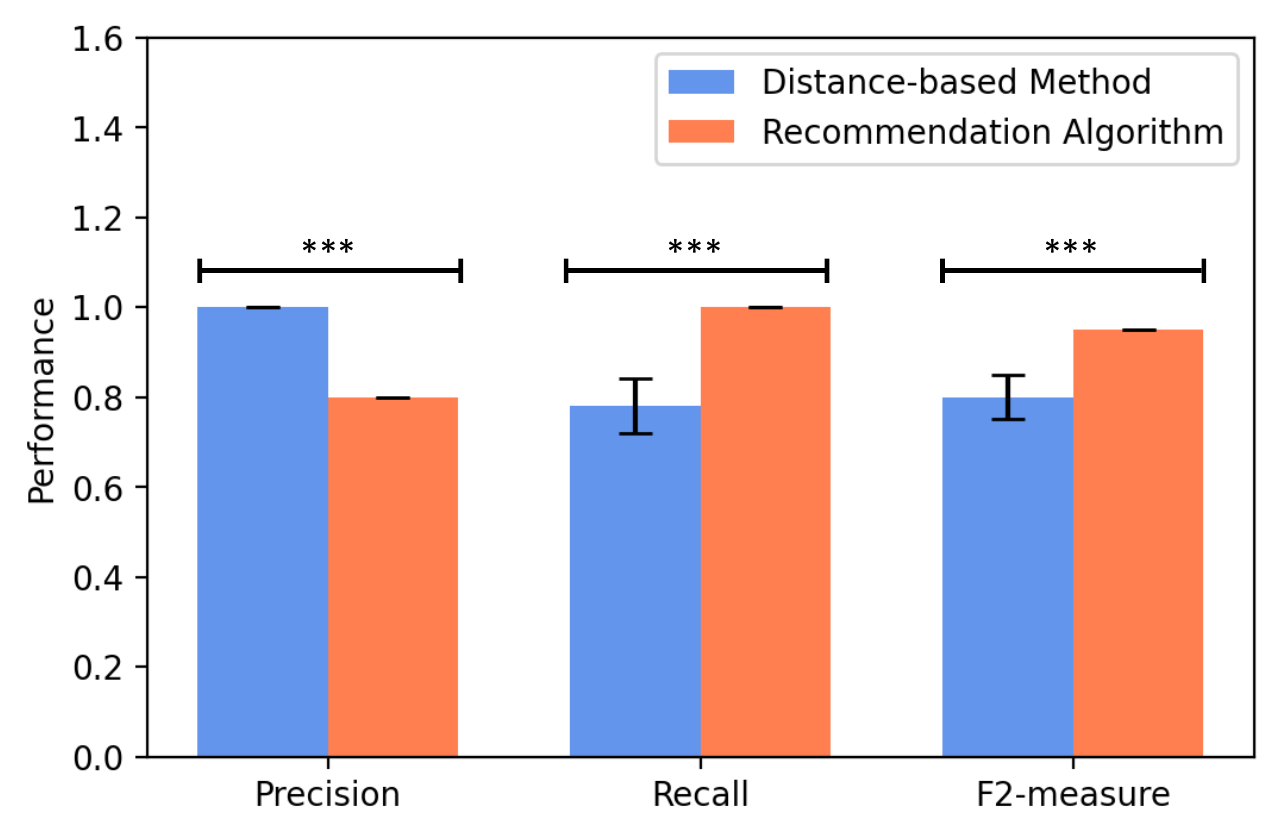}
    \caption{Hurricane Ian (2022)}
  \end{subfigure}
  \vspace{-4pt}
  \caption{Precision, recall, and $F_2$ scores for identifying affected
  ports across the sampled landfall scenarios.}
  \label{fig:classification}
\end{figure}

At the retrieval stage, the recommendation algorithm selects the ten ports nearest to each sampled landfall location, while the distance-based method selects all ports within 400~km of that location. Figure~\ref{fig:classification} compares how well these two methods identify the ports that were actually affected. FM, GBDT+LR, and WDL share the same retrieval step and therefore have identical classification results. The recommendation algorithm identifies all four affected ports during Hurricane Nicole and all eight during Hurricane Ian, achieving full recall in both cases. With ten ports selected, its precision is 0.40 for Nicole and 0.80 for Ian. The distance-based method achieves higher precision but lower recall for both hurricanes. The mean $F_2$ scores are similar for Nicole, with a slight advantage for the distance-based method, while the recommendation algorithm achieves a higher mean $F_2$ score for Ian.

\vspace{-16pt}

\subsection{Comparison of Ranking Performance}

Across both test hurricanes, WDL provides the closest agreement with the observed port impact rankings, and its ranking performance varies less across landfall scenarios than that of the distance-based baseline (Table~\ref{tab:model_performance}). For Hurricane Nicole, mean AO increases from 0.454 for the baseline to 0.842 for WDL, while the standard deviation decreases from 0.112 to 0.025. The same pattern holds for Hurricane Ian, where mean AO increases from 0.607 to 0.944 and the standard deviation decreases from 0.090 to 0.011. WDL also achieves the lowest MAE and RMSE for both hurricanes, with mean absolute errors of 0.477 days for Nicole and 0.871 days for Ian. Among the other two learned models, GBDT+LR estimates impact days more accurately than FM, while FM produces rankings that agree more closely with the observed order.

\begin{table}[h]
\centering
\caption{Impact-day estimation and port ranking performance across
20 landfall scenarios for each test hurricane.}
\label{tab:model_performance}
\renewcommand{\arraystretch}{0.95}
\setlength{\tabcolsep}{8pt}
\small
\begin{tabular}{lccc}
\toprule
\textbf{Model}
& \textbf{MAE (days)} $\downarrow$
& \textbf{RMSE (days)} $\downarrow$
& \textbf{AO} $\uparrow$ \\
\midrule
\multicolumn{4}{l}{\textbf{Hurricane Nicole (H1)}} \\
Distance-based & -- & -- & 0.454 (0.112) \\
FM      & 1.346 (0.048) & 1.743 (0.073) & 0.682 (0.027) \\
GBDT+LR & 0.498 (0.013) & 0.802 (0.012) & 0.440 (0.050) \\
WDL     & \textbf{0.477} (0.028) & \textbf{0.691} (0.021)
        & \textbf{0.842} (0.025) \\
\addlinespace[3pt]
\multicolumn{4}{l}{\textbf{Hurricane Ian (H5)}} \\
Distance-based & -- & -- & 0.607 (0.090) \\
FM      & 1.004 (0.088) & 1.301 (0.111) & 0.723 (0.023) \\
GBDT+LR & 0.915 (0.001) & 1.180 (0.002) & 0.682 (0.001) \\
WDL     & \textbf{0.871} (0.036) & \textbf{0.933} (0.020)
        & \textbf{0.944} (0.011) \\
\bottomrule
\end{tabular}

\medskip
\begin{minipage}{0.94\linewidth}
\footnotesize
\textit{Note:} Values are means across landfall scenarios, with standard
deviations in parentheses. Arrows indicate whether lower ($\downarrow$)
or higher ($\uparrow$) values are preferred. Bold denotes the best mean
for each metric and hurricane.
\end{minipage}
\end{table}

For inspection and recovery planning, a practical question is whether the ports with the longest observed disruptions remain near the top of the estimated ranking as the assumed landfall location changes. In the three scenarios shown for Nicole, WDL consistently ranks Miami first and Palm Beach second, matching the observed order. The distance-based baseline places Miami seventh, third, and sixth, respectively (Table~\ref{tab:port_ranking_nicole}). For Ian, WDL reproduces the observed order of the four highest-ranked ports across all three scenarios: Everglades, Miami, Manatee, and Tampa. The baseline places either Miami or Manatee first (Table~\ref{tab:port_ranking_ian}). Agreement is less consistent farther down the list. For example, WDL ranks Everglades sixth in all three Nicole scenarios, although it was third in the observed ranking. In these two cases, the clearest benefit of incorporating port and cyclone characteristics is the ability to identify the ports facing the longest disruptions even as the sampled landfall location changes.

\begin{table}[htbp]
\centering
\caption{Observed and estimated port impact rankings for three
landfall scenarios under Hurricane Nicole.}
\label{tab:port_ranking_nicole}
\setlength{\tabcolsep}{3.5pt}
\renewcommand{\arraystretch}{1.0}
\resizebox{\textwidth}{!}{%
\begin{tabular}{cccccccc}
\toprule
\textbf{Rank} & \textbf{Observed} &
\textbf{Dist-1} & \textbf{Dist-2} & \textbf{Dist-3} &
\textbf{WDL-1} & \textbf{WDL-2} & \textbf{WDL-3} \\
\midrule
1  & Miami        & Canaveral    & Palm Beach   & Canaveral
   & Miami        & Miami        & Miami \\
2  & Palm Beach   & Palm Beach   & Everglades   & Palm Beach
   & Palm Beach   & Palm Beach   & Palm Beach \\
3  & Everglades   & Tampa        & Miami        & Everglades
   & Manatee      & Manatee      & Manatee \\
4  & Jacksonville & Manatee      & Canaveral    & Tampa
   & Jacksonville & Mobile       & Jacksonville \\
5  & --           & Everglades   & Manatee      & Manatee
   & Mobile       & Jacksonville & Mobile \\
6  & --           & Jacksonville & Tampa        & Miami
   & Everglades   & Everglades   & Everglades \\
7  & --           & Miami        & --           & Jacksonville
   & Pascagoula   & Pascagoula   & Pascagoula \\
8  & --           & --           & --           & --
   & Panama       & Panama       & Panama \\
9  & --           & --           & --           & --
   & Canaveral    & Canaveral    & Canaveral \\
10 & --           & --           & --           & --
   & Tampa        & Tampa        & Tampa \\
\bottomrule
\end{tabular}}

\medskip
\begin{minipage}{\linewidth}
\footnotesize
\textit{Note:} Dist and WDL denote the distance-based baseline and the
WDL ranking model. Matching suffixes refer to the same sampled landfall
location. A dash indicates that the list contains no port at that rank.
\end{minipage}
\end{table}

\begin{table}[htbp]
\centering
\caption{Observed and estimated port impact rankings for three
landfall scenarios under Hurricane Ian.}
\label{tab:port_ranking_ian}
\setlength{\tabcolsep}{3.5pt}
\renewcommand{\arraystretch}{1.0}
\resizebox{\textwidth}{!}{%
\begin{tabular}{cccccccc}
\toprule
\textbf{Rank} & \textbf{Observed} &
\textbf{Dist-1} & \textbf{Dist-2} & \textbf{Dist-3} &
\textbf{WDL-1} & \textbf{WDL-2} & \textbf{WDL-3} \\
\midrule
1  & Everglades   & Miami        & Manatee      & Manatee
   & Everglades   & Everglades   & Everglades \\
2  & Miami        & Everglades   & Tampa        & Tampa
   & Miami        & Miami        & Miami \\
3  & Manatee      & Palm Beach   & Everglades   & Everglades
   & Manatee      & Manatee      & Manatee \\
4  & Tampa        & Manatee      & Miami        & Palm Beach
   & Tampa        & Tampa        & Tampa \\
5  & Palm Beach   & Tampa        & Palm Beach   & Canaveral
   & Palm Beach   & Canaveral    & Jacksonville \\
6  & Jacksonville & Canaveral    & Canaveral    & --
   & Canaveral    & Palm Beach   & Palm Beach \\
7  & Panama       & --           & --           & --
   & Jacksonville & Jacksonville & Canaveral \\
8  & Canaveral    & --           & --           & --
   & Pascagoula   & Pascagoula   & Panama \\
9  & --           & --           & --           & --
   & Mobile       & Mobile       & Pascagoula \\
10 & --           & --           & --           & --
   & Panama       & Panama       & Mobile \\
\bottomrule
\end{tabular}}

\medskip
\begin{minipage}{\linewidth}
\footnotesize
\textit{Note:} Dist and WDL denote the distance-based baseline and the
WDL ranking model. Matching suffixes refer to the same sampled landfall
location. A dash indicates that the list contains no port at that rank.
\end{minipage}
\end{table}

\section{Conclusion}

When a single TC threatens several ports, two questions arise for preparedness planning: which ports are likely to be affected, and where are disruptions likely to last longest? This study addresses these questions by formulating the assessment of relative impacts across ports as a recommendation problem. Using AIS-derived impact days for 37 ports and 31 TCs, the framework draws on historical interactions involving ports and cyclones with similar characteristics when observations at an individual port are limited. For each plausible landfall location, it first retrieves potentially affected ports and then ranks them by their estimated number of impact days. Within the ranking stage, we compare three models that differ in how they relate combinations of port and cyclone characteristics to impact duration. FM represents pairwise feature interactions through latent vectors, GBDT+LR uses combinations of conditions along tree paths as inputs to linear regression, and WDL combines explicit feature crosses with learned embeddings. Comparing these models on the same candidate ports allows us to examine how well each representation captures differences in impact duration across ports exposed to the same TC.

In tests on Hurricanes Nicole and Ian, the retrieval stage identifies all four ports with observed impact days during Nicole and all eight during Ian across the 20 sampled landfall locations for each hurricane. Among the three ranking models, WDL achieves the lowest MAE and RMSE and the closest agreement with the observed port impact rankings. Relative to the distance-based baseline, its mean ranking agreement is 85.5\% higher for Nicole and 55.5\% higher for Ian, with less variation across landfall scenarios. The individual port rankings illustrate what this improvement means for preparedness planning. In the three scenarios shown for Nicole, the distance-based baseline ranks Miami between third and seventh as the sampled landfall location changes. WDL consistently places Miami first and Palm Beach second, matching the two ports with the longest observed disruptions. For Ian, WDL reproduces the observed order of the four highest-ranked ports in all three illustrated scenarios. These cases suggest that, once potentially affected ports have been identified, accounting for port and cyclone characteristics can help distinguish where disruptions are likely to last longest. Port authorities could use that distinction to prioritize post-cyclone inspections and allocate recovery resources across ports threatened by the same event.

This study represents landfall uncertainty by sampling plausible locations along the coastline within 120~km of the recorded landfall point. Future work could extend this analysis using successive forecast advisories to examine how early the ports facing the longest disruptions can be identified and how their rankings change as the forecast track and intensity evolve. Evaluating additional TCs and ports across a wider geographic area would also help establish how well the findings from Nicole and Ian transfer to other events and coastal regions.

\bibliographystyle{elsarticle-harv}  
\bibliography{cas-refs}

\appendix
\section{Model Implementation Details}
\label{app:implementation}

\paragraph{Input representation and training protocol.}
Each example corresponds to a port--cyclone pair under a sampled landfall location. The regression target is the observed number of impact days for that pair, which remains fixed across landfall scenarios. The archived preprocessing pipeline discretizes annual port tonnage at 10, 50, and 100 million tons. Distances from the port to the cyclone formation point and to the landfall location are discretized at 100, 200, 300, 400, and 500~km. Landfall wind speed is grouped using thresholds of 64, 82, 95, 112, and 136 knots, and the time from formation to landfall is rounded to the nearest day. The resulting variables and the remaining categorical port and cyclone attributes are encoded as categorical fields.

We split the data by tropical cyclone, keeping all observations from a given TC in the same partition. Approximately 20\% of the cyclones from 2017, 2018, 2019, and 2021 are reserved for validation, and the remainder are used for training. Hurricanes Nicole and Ian in 2022 are held out for testing.

\paragraph{FM and GBDT+LR implementations.}
The archived FM implementation uses PyFM with 100 latent factors, 100 training iterations, an initial learning rate of 0.001, and the \texttt{optimal} learning-rate schedule. The one-hot encoded features are stored in a compressed sparse row matrix, and FM is fitted to the observed number of impact days using squared-error loss.

The archived GBDT+LR implementation uses \texttt{GradientBoostingClassifier} with 100 boosting iterations, a maximum tree depth of 30, a learning rate of 0.05, and \texttt{random\_state=16}. At each boosting iteration, the implementation selects the first class-specific tree and records the leaf index reached by each example. The leaf indices from the selected trees are one-hot encoded and used as inputs to \texttt{LinearRegression} to estimate impact days. The leaf encoder is fitted on the training examples and reused for validation and testing.

\paragraph{WDL reference implementation.}
The WDL reference configuration uses a wide linear component and a deep component with an eight-dimensional embedding for each categorical field. The wide component takes the original one-hot features and their pairwise categorical crosses as inputs. Feature vocabularies are constructed from the training data, with a reserved index for unseen categories. In the deep component, the embeddings are concatenated and passed through fully connected layers with 256, 128, and 64 units. Each layer uses ReLU activation followed by dropout with probability 0.2. A linear output combines the two components to estimate the number of impact days.

For $N$ training examples, the reference configuration uses the following objective:
\begin{equation}
\mathcal{L}
=
\frac{1}{N}\sum_{i=1}^{N}(y_i-\hat{y}_i)^2
+
10^{-3}\lVert w_{\mathrm{wide}}\rVert_2^2
+
10^{-5}\lVert W_{\mathrm{deep}}\rVert_2^2,
\label{eq:wdl_reference_loss}
\end{equation}
where $y_i$ and $\hat{y}_i$ are the observed and estimated numbers of impact days. The regularization terms sum the squared weights in the wide and deep components, respectively, excluding biases.

The reference model is trained with Adam at a learning rate of $10^{-3}$, a batch size of 256, and 100 epochs. The checkpoint with the lowest validation mean squared error is retained. The random seeds for Python, NumPy, and PyTorch are set to 42. The reference implementation saves the feature mappings, validation assignments, dependency versions, and model checkpoints.

\end{document}